\documentclass[Afour,times,sageh]{sagej}
\pdfoutput=1
\usepackage{moreverb,url}

\usepackage[colorlinks,bookmarksopen,bookmarksnumbered,citecolor=red,urlcolor=red]{hyperref}
\usepackage{amsmath}
\usepackage{makecell}
\newcommand\BibTeX{{\rmfamily B\kern-.05em \textsc{i\kern-.025em b}\kern-.08em
T\kern-.1667em\lower.7ex\hbox{E}\kern-.125emX}}

\newcommand{\eg}{\textit{e}.\textit{g}.}

\newcommand{\argmax}{\mathop{\rm arg~max}\limits}

\def\volumeyear{2016}

\begin{document}

\runninghead{et al.}

\title{Understanding hand-object manipulation by modeling the contextual relationship between actions, grasp types and object attributes}

\author{Minjie Cai\affilnum{1},
Kris M. Kitani\affilnum{2} and
Yoichi Sato\affilnum{1}}

\affiliation{\affilnum{1}Institute of Industrial Science, The University of Tokyo, Japan\\
\affilnum{2}Robotics Institute, Carnegie Mellon University, USA}

\corrauth{Minjie Cai, Institute of Industrial Science, The University of Tokyo,
Komaba 4-6-1, Meguro-ku,
Tokyo 153-8505, Japan.}

\email{cai-mj@iis.u-tokyo.ac.jp}

\begin{abstract}
This paper proposes a novel method for understanding daily hand-object manipulation by developing computer vision-based techniques. Specifically, we focus on recognizing hand grasp types, object attributes and manipulation actions within an unified framework by exploring their contextual relationships. Our hypothesis is that it is necessary to jointly model hands, objects and actions in order to accurately recognize multiple tasks that are correlated to each other in hand-object manipulation. In the proposed model, we explore various semantic relationships between actions, grasp types and object attributes, and show how the context can be used to boost the recognition of each component. We also explore the spatial relationship between the hand and object in order to detect the manipulated object from hand in cluttered environment. Experiment results on all three recognition tasks show that our proposed method outperforms traditional appearance-based methods which are not designed to take into account contextual relationships involved in hand-object manipulation. The visualization and generalizability study of the learned context further supports our hypothesis.
\end{abstract}


\maketitle

\section{Introduction}

The understanding of hand-object manipulation (HOM) is concerned with how the hand interacts with the object under certain purpose, requiring information such as the poses of a hand for holding/manipulating an object, the attribute information of the manipulated object and the undergoing action. The ability to understand HOM activities automatically from visual sensing is important for the robotics community with potential applications such as robotic hand design and robotic action learning. In robotic hand design, the study of hand grasping behavior in daily manipulation tasks provides critical information about hand functions that can be used for robotic hand development (\cite{cutkosky1989grasp,zheng2011investigation,bullock2013finding,dollar2014classifying}). It can also facilitate robotic action learning by studying the relationship between different components (hands, objects and actions) in performing a manipulation task (\cite{feix2014analysis,yang2015robot}). 

%Wearable cameras enable continuous recording of unconstrained natural hand-object interactions at a large scale, both in time and space, and provides an ideal first-person point-of-view under which hands and objects are often visible up-close in the visual field. In this work, we develop automatic egocentric (first-person) vision techniques that can be used as a tool to promote the studies of hand manipulation in real-life settings.

In this work, we aim to recognize grasp types, object attributes and actions of manipulation tasks recorded with a wearable camera. The three terms provide very useful information for understanding HOM activities, and are defined as follows: \textit{Grasp types} are a discrete set of canonical hand poses often used to describe various strategies for holding objects stably in hand. For example, the use of all fingers around a curved object like a cup is called a medium wrap. \textit{Object attributes} characterize physical properties of the objects such as rigidity and shape. \textit{Actions} in this work refer to different patterns of hand-object interactions such as open or pour. A wearable camera enables continuous recording of unconstrained hand-object interactions and offers an ideal viewing perspective for recording and studying manipulation activities in real-life settings. However, the recognition of real-world HOM activities is challenging. There are often occlusions of a hand, especially fingers, during hand-object interactions, making it hard to observe and recognize hand grasps. It is also challenging to detect the manipulated object and infer its attribute information in cluttered background. Although recent years have seen the popularity of deep learning techniques in various vision recognition tasks, their performance is often limited with insufficient training data and the correlation between different recognition tasks of hands, objects and actions are not considered.

In this work, we propose an unified model (Figure~\ref{fig_graphical_model}) in which the grasp types, object attributes and actions are recognized together by exploring their contextual relationships. Our hypothesis is that the actions, hands and objects are strongly correlated with each other in HOM activities, and thereby it is important to jointly model them together. Jointly modeling hand grasps, objects and actions has its empirical basis in neuroscience and psychology (\cite{kjellstrom2008visual,feix2014analysis1}), and our work firstly presents a computational way to explicitly modeling these relationships. Specifically, three kinds of contextual relationships are studied in the proposed model. 
The \textit{functional context} models the functional constraints on grasp types and object attributes within different manipulation actions.
The \textit{physical context} models the semantic relationship between hand grasp types and attributes of the grasped object under physical constraints.
The \textit{spatial context} models the spatial relationship between the hand and the manipulated object.
With these contextual relationships serving as high-level constraints, we utilize deep Convolutional Neural Networks (CNNs) to model the visual evidence for grasp types, object attributes and actions respectively. In particular, we develop a CNN to detect the hand and the grasped object simultaneously based on spatial hand-object relationship. 
We show in experiments that our method significantly improves the recognition performance of all three tasks compared with previous approaches which did not consider mutual context involved in these tasks. Generalizability study further indicates that the contextual relationships are consistent across different datasets and therefore could be exploited as common knowledge in understanding hand-object manipulation.

%Object attributes (\eg, thick or long shape of a bottle) have strong impacts on the selection of hand grasp types (\eg, \textit{Large Wrap}). Thus, with the knowledge of object attributes, we obtain a prior information about the possibility of different grasp types. On the other hand, knowing the grasp type also reveals attributes of the object being manipulated.
%Grasp types together with object attributes provide complementary information for characterizing the hand action. There are several advantages for jointly modeling actions in this way: (1) Grasp type helps describe the functionality of an action, whether it requires more power, or more flexible finger coordination; (2) Object attributes provide a general description about the manipulated object and indicates possible interaction patterns; (3) Semantic information of grasp types and object attributes enable the model to encode high-level constraints, and as a result, the learned action model is immediately interpretable.

Overall, this paper extends our previous work that was published in \cite{cai2016understanding}. Novel contributions over our previous publication are summarized as follows:
\begin{itemize}
  \item An unified model is constructed for studying various contextual relationships between hands, objects and actions together (Section~\ref{s_architecture}).
  \item A detection CNN is developed to detect hand and object simultaneously (Section~\ref{ssec_train_detector}).
  \item An iterative inference method is developed to optimize multiple tasks together (Section~\ref{sec_inference}).
  \item More detailed visualization and analysis of the experiment results is provided (Section~\ref{s_experiment}).
\end{itemize}

The rest of the paper is organized as follows. Section~\ref{s_related_work} presents related work. The architecture and each component of our proposed model are described in Section~\ref{s_architecture}. Section~\ref{s_learning} and Section~\ref{sec_inference} elaborate the details of learning and inference of the model. Experiment results are given in Section~\ref{s_experiment}. Section~\ref{s_conclusion} concludes the paper and discusses future work.

\section{Related Work}
\label{s_related_work}

%\subsection{Hand grasp}

\textit{Hand grasp} has been studied for decades to better understand the use of human hands (\cite{napier1956prehensile,santello1998postural,romero2013non,bullock2013grasp,huang2015we}). Grasp taxonomies have also been proposed to facilitate hand grasp analysis (\cite{cutkosky1989grasp, kang1993toward, feix2009comprehensive, liu2014taxonomy, feix2016grasp}). 
Approaches for vision-based hand grasp analysis were developed primarily in structured environment where hand-object interactions are well recorded by camera arrays or depth sensors (\cite{kjellstrom2008visual, hamer2009tracking, oikonomidis2011full, romero2013non}). 
\cite{cai2015scalable} first developed appearance-based method to recognize hand grasp types from manipulation tasks recorded in real-world environment with a wearable RGB camera. \cite{saran2015hand} then used detected hand parts as intermediate representation to distinguish between fine-grained grasp types. \cite{yang2015grasp} and \cite{rogez2015understanding} further utilized deep learning techniques to improve grasp recognition performance. Nevertheless, hand grasp recognition on 2D images still remains a challenging problem due to limited view of hand appearance and visual ambiguity between different grasp types (\cite{cai2017ego}).

%\subsection{Attribute classification}

\textit{Visual attributes} (physical properties inferred from image appearance) are often used as intermediate representation for many applications, such as object recognition~(\cite{farhadi2009describing, lampert2009learning, vedaldi2014understanding}), facial verification~(\cite{kumar2009attribute}), image retrieval and tagging (\cite{siddiquie2011image, parikh2011relative, zhang2014panda}). \cite{lampert2009learning} performs object detection based on a human-specified high-level description of the target classes for which no training examples are available. The description consists of attributes like shape, color or even geographic information. \cite{parikh2011relative} explored the relative strength of attributes by learning a rank function for each attribute which can be used to generate richer textual descriptions. In this work, we extract attribute information from the manipulated object and use it as context to improve the recognition performance of grasp types and actions.

The correlations between objects and hand grasps in hand-object interactions have been widely studied in neuroscience and psychology. It has been shown that humans use the same or similar grasp types for certain types of objects, and the shape of the object has a large influence on the applied grasp (\cite{klatzky1987knowledge, gilster2012contact}). Recently, \cite{feix2014analysis} investigated the relationship between grasp types and object attributes in a large real-world human grasping dateset by manual inspection. In our prior work (\cite{cai2016understanding}), a Bayesian network is trained to automatically learn the contextual relationship between grasp types and object attributes. The context information has then been used to improve the recognition performance of both grasp types and object attributes. In this paper, we extend our prior work to incorporate the context from actions in a more complete model.

%\subsection{Manipulation action}

Recognizing actions of hand-object interactions under first-person vision paradigm has become attractive in recent years since a wearable camera provides an ideal viewing perspective for recording and analyzing hand-object interactions. \cite{fathi2011understanding, fathi2012learning} used appearance around the regions of hand-object interactions to recognize egocentric actions. \cite{pirsiavash2012detecting} has shown that detecting objects in the scene helps to infer daily hand activities. \cite{li2015delving} performed a systematic evaluation of different features and provided a list of best practices of combining different cues for activity recognition. \cite{ma2016going} further proposed a deep learning architecture for egocentric action recognition, which integrates both appearance and motion information. 
However, previous work only treated hand activity recognition as an image or video classification problem without full understanding of the semantic relationships involved. We argue that it is important to obtain a detailed understanding of the interactions between the hands and the objects in HOM activities.
Our work is inspired by a number of works that studied the interactions between human poses and objects in human activities (\cite{singh2010multiple, yao2010modeling, sadeghi2011recognition}). Different from these work, we aim to better understand HOM activities by modeling the contextual relationships between actions, hand grasp types and object attributes.

\section{Architecture}
\label{s_architecture}

Our goal is to recognize the hand grasp type, attribute information of the manipulated object, and the manipulation action within a HOM activity.
Based on our hypothesis that hands, objects and actions are correlated with each other in HOM activities, we propose a novel model to jointly infer grasp types, object attributes and actions by exploring their contextual relationships in an unified framework.

\begin{figure}[t]
    \centering
    \includegraphics[width=\linewidth]{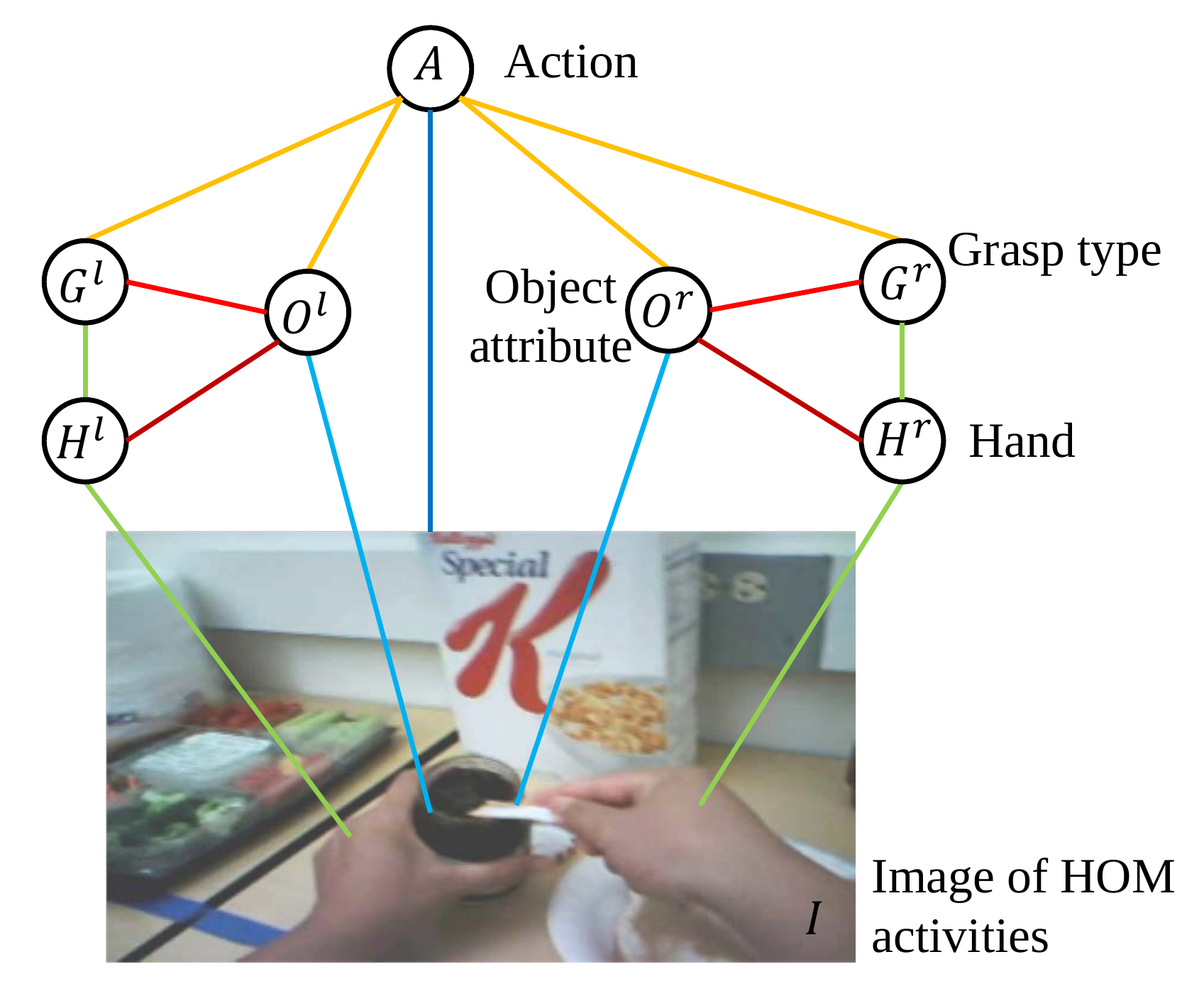}
    \caption{An illustration of our proposed graphical model. $A$ denotes the hand action class, $G$ the grasp type class, $O$ the object attribute class, and $P$ the spatial positions of hands and objects. We define $G$ and $O$ for both two hands. The edges denoted by different colors correspond to different components of our model.}
    \label{fig_graphical_model}
\end{figure}

Figure~\ref{fig_graphical_model} illustrates the proposed model. The model can be considered as a conditional random field. Given an image (or image sequence) $I$ of HOM activities, we jointly model the action class $A$, the hands $H=\{H^l,H^r\}$ with grasp types $G=\{G^l,G^r\}$, and the attributes of the manipulated objects $O=\{O^l,O^r\}$. $l,r$ indicate the left and right side respectively.
The proposed model could be understood functionally as two different parts: The first part models different contextual relationships between hands, objects and actions. The second part models visual evidence of grasp types, object attributes, and actions conditioned on image appearance.
Overall, the model is represented as
\begin{multline}
\label{equation_potential}
\Psi(A,G,O,H|I)=\\
\Psi_{FC}(A,G,O)+\Psi_{PC}(G,O)+\Psi_{SC}(H,O)+\\
\Psi_G(G,H|I)+\Psi_O(O|I)+\Psi_A(A|I),
\end{multline}
where $\Psi_{FC}$ models the semantic relationship between actions, grasp types and object attributes under functional constraints, $\Psi_{PC}$ models the semantic relationship between grasp types and object attributes under physical constraints, $\Psi_{SC}$ models the spatial relationship between the hand and the grasped object, $\Psi_G,\Psi_O,\Psi_A$ models the visual evidence based on discriminative classifiers trained for grasp types, object attributes and actions respectively. Next, we explain the details of each component.

\subsection{Functional context}
\label{ssec_functional_context}

We first consider the functional context denoted as $\Psi_{FC}(A,G,O)$ which models the contextual relationship between actions, grasp types and object attributes under functional constraints.  
It is observed that grasp types together with object attributes provide complementary characterization for the functionality of different manipulation actions. For example, given a pen, the grasp type of \textit{writing tripod} reveals the functionality of action \textit{write}, while the grasp type of \textit{thumb-2 finger} reveals the functionality of action \textit{take}. On the other hand, the knowledge of the action being performed provides functional constraints on the hand grasp type and the attributes of the manipulated object. For example, the action of \textit{scoop} indicates the high probability of combination of a container (such as a bottle) held by some power grasp and a long-shape tool (such as a spoon) held by some precision grasp. Therefore, $\Psi_{FC}(A,G,O)$ is parameterized as
\begin{multline}
\label{equation_ago}
\Psi_{FC}(A,G,O)=\sum_{k=1}^{N_a}\ \sum_{i,j=0}^{N_g}\sum_{m,n=0}^{N_o}\textbf{1}_{(A=a_k)}\cdot\textbf{1}_{(G^l=g_i)}\\
\cdot\textbf{1}_{(G^r=g_j)}\cdot\textbf{1}_{(O^l=o_m)}\cdot\textbf{1}_{(O^r=o_n)}\cdot\alpha_{k,i,j,m,n},
\end{multline}
where $N_a$ is the number of action classes and $a_k$ denotes the $k$-th action class. $N_g$ is the number of grasp types and $g_{i/j}$ denotes the $i/j$-th grasp type ($i/j=0$ means no hand). $N_o$ is the number of object attributes and $o_{m/n}$ denotes the $m/n$-th object attribute ($m/n=0$ means no object). $1_{(\cdot)}$ is an indicator function and has value $1$ when the condition inside $(\cdot)$ is fulfilled, otherwise $0$. $\alpha_{k,i,j,m,n}$ indicates the strength of functional compatibility between the action $a_k$, the grasp type $g_i, g_j$ and object attribute $o_m, o_n$ at the left hand and the right hand ($l,r$ indicate the side).

\subsection{Physical context}

We also model the contextual relationship between grasp types and object attributes under physical constraints as denoted by $\Psi_{PC}(G,O)$. Object attributes pose physical constraints on the affordance of different grasp types. For example, a bottle cap with a round shape is more likely to be held with the grasp type of \textit{precision sphere} than with other grasp types. Hence, with the knowledge of object attributes, we obtain a prior information about the possibility of different grasp types. On the other hand, knowing the grasp type also reveals attributes of the grasped object. Therefore, $\Psi_{PC}(G,O)$ is parameterized as
\begin{equation}
\label{equation_go}
\Psi_{PC}(G,O)=\sum_{s\in\{l,r\}}\sum_{i=1}^{N_g}\sum_{m=1}^{N_o}\textbf{1}_{(G^s=g_i)}\cdot\textbf{1}_{(O^s=o_{m})}\cdot\beta_{i,m},
\end{equation}
where $s$ indicates the side of hand. $N_g, N_o, g_i, o_m$ have the same meaning as described in Section~\ref{ssec_functional_context}. $\beta_{i,m}$ indicates the strength of physical constraints between grasp type $g_i$ and object attribute $o_m$.

\subsection{Spatial context}
\label{ss_spatial_context}

In addition to the above semantic relationships, we also explore the spatial context between the hand and grasped object denoted as $\Psi_SC(H,O)$. Unlike the hand that has unique skin color and consistent size, object has inconsistent appearance and is often partially occluded by the hand during manipulation. Hence, the task of object detection is more challenging than hand detection. In this work we propose to explore the spatial context of hand for estimating the location of the grasped object. We observe that hand appearance provides important hint about the relative location and size of the grasped part of the object (not the whole object, and we will refer to ``grasped part of the object'' simply as ``object'' in the rest of paper). For instance, object size is related to hand opening and relative object location is consistent with hand orientation. The spatial context of $\Psi_SC(H,O)$ is represented as
\begin{equation}
\label{equation_spatial}
\Psi_{SC}(H,O)=\sum_{s\in\{l,r\}}\gamma_s\cdot p(P_o^s|P_{h\to o}^s),
\end{equation}
where $p(P_o^s|P_{h->o}^s)$ is the likelihood of observing $P_o^s$, the object location, given the reference object location $P_{h\to o}^l$ estimated from the $s$-side hand region, and is computed by intersection over union of the two object locations (bounding boxes). $\gamma_s$ is the weight for the spatial relationship between the hand and the grasped object.

\subsection{Grasp types}

$\Psi_G(G,H,I)$ models visual evidence of grasp types. It is further composed by two components. The first component models the likelihood of detecting the hand $H$ in image $I$. The second components models the visual belief of the grasp type $G$ conditioned on image appearance within $H$. $\Psi_G(G,H,I)$ is represented as
\begin{multline}
\label{equation_grasp}
\Psi_{G}(G,H|I)=\\
\sum_{s\in\{l,r\}}\Big(\zeta_s^T\cdot \phi_{h}(P_h^s|I)+\sum_{i=1}^{N_g}1_{(G^s=g_i)}\cdot\eta_{i}^T\cdot \phi_{g}(P_h^s)\Big),
\end{multline}
where $\phi_{h}(P_h^s|I)$ is the vector of hand detection scores obtained from a hand detector, and $\phi_{g}(P_h^s)$ is the vector of classification scores obtained from a grasp type classifier. $\zeta_s$ is the set of weights for the detection scores corresponding to the $s$-side hand, and $\eta_{i}$ is the set of weights for the classification scores corresponding to grasp type $g_i$.

\subsection{Object attributes}

Similar to grasp types, $\Psi_O(O,I)$ models visual evidence for object attributes. However, the object location is not directly modeled from image appearance (it is modeled by spatial context in Section~\ref{ss_spatial_context}). $\Psi_O(O,I)$ is represented as
\begin{equation}
\label{equation_attribute}
\Psi_{O}(O|I)=\sum_{s\in\{l,r\}}\sum_{m=1}^{N_o}\textbf{1}_{(O^s=o_m)}\cdot \lambda_{m}^T\cdot\phi_{o}(P_o^s),
\end{equation}
where $\phi_{o}(P_o^s)$ is the output of object attribute classifier obtained from the object region in $P_o^s$. $\lambda_{m}$ encodes the set of weights for the classification scores corresponding to object attribute $o_m$.

\subsection{Actions}

$\Psi_A(A,I)$ models visual evidence for manipulation actions. Different from grasp types and object attributes which are recognized from image regions after a detection procedure, actions are recognized directly from a whole image. $\Psi_A(A,I)$ is represented as
\begin{equation}
\label{equation_action}
\Psi_{A}(A|I)=\sum_{i=1}^{N_a}1_{(A=a_i)}\cdot\xi_{i}^T\cdot \phi_{a}(I),
\end{equation}
where $\phi_{a}(I)$ is the output vector of an action classifier. $\xi_{i}$ is the set of weights corresponding to action class $a_i$.

\section{Learning}
\label{s_learning}

In this section, we describe the approaches for learning different components of our model. We first introduce the class definition of grasp types and object attributes in Section~\ref{ssec_class} that are important components of the model. We then show how to train hand and object detector in Section~\ref{ssec_train_detector}, and train classifiers for grasp types, object attributes and actions in Section~\ref{ssec_train_classifier}. We explain how to learn the model parameters in Section~\ref{ssec_train_parameter}. Implementation details about the utilized CNNs are given in Section~\ref{ssec_implementation}.

\subsection{Class definition}
\label{ssec_class}

\subsubsection{Grasp types}
\label{sssec_grasp_type}

\begin{figure}[t]
    \centering
    \includegraphics[width=0.95\linewidth]{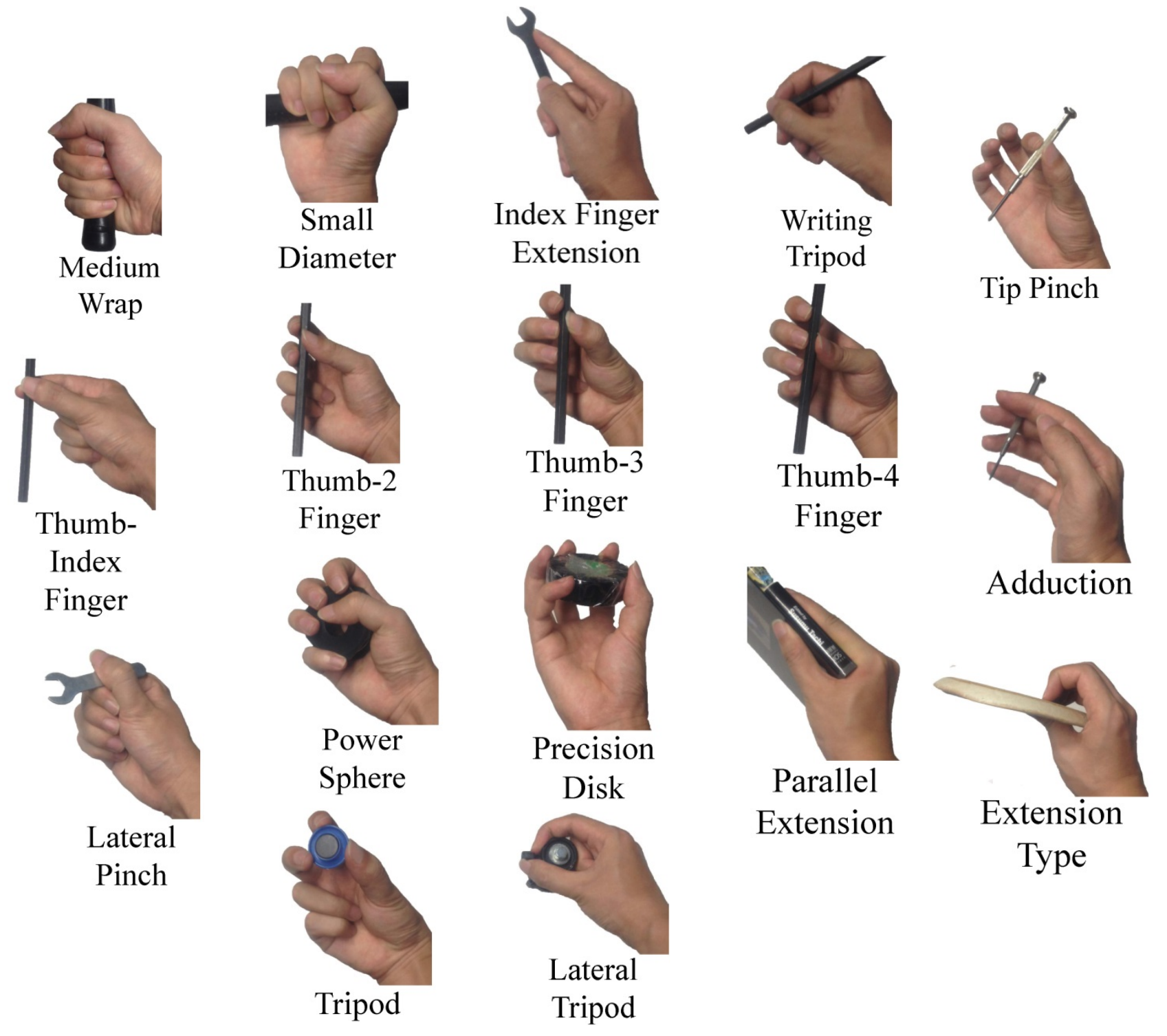}
    \caption{Examples of hand grasp types from a commonly used grasp taxonomy defined in \cite{feix2009comprehensive}. Images are adapted from \cite{cai2015scalable}.}
    \label{fig_grasp_list}
\end{figure}

Hand grasp types are important for understanding HOM activities since they characterize how hands hold the objects, as well as the action functionality to be implemented. A number of work (\cite{cutkosky1989grasp,feix2016grasp}) have investigated the categorization of hand poses when holding an object into a discrete set of types to facilitate the study of hand grasps. The resulting grasp taxonomies are created according to some classification criterion mainly based on functionality, object shape, and finger articulation. A commonly used grasp taxonomy is illustrated in Figure~\ref{fig_grasp_list}. However, it is not an appropriate way to use the full sized taxonomy for any specific tasks as the proportion of different grasp types varies a lot. It also has been shown in \cite{cai2017ego} that there is a trade-off between the size of grasp taxonomy and the classification accuracy. In this work, we adopt an unsupervised clustering method of \cite{huang2015we} to first discover a diverse set of hand grasps based on hand appearance from the dataset being used. Then the obtained clusters of hand images are examined with existing grasp taxonomy to define the set of grasp types to be used in the current task. We assign new grasp type to those clusters whose hand poses are undefined in existing grasp taxonomies, and discard the clusters with small member size. The details of the used grasp types are described later in experiments.

\subsubsection{Object attributes}
\label{sssec_object_attribute}

\begin{figure}[t]
    \centering
    \includegraphics[width=0.9\linewidth]{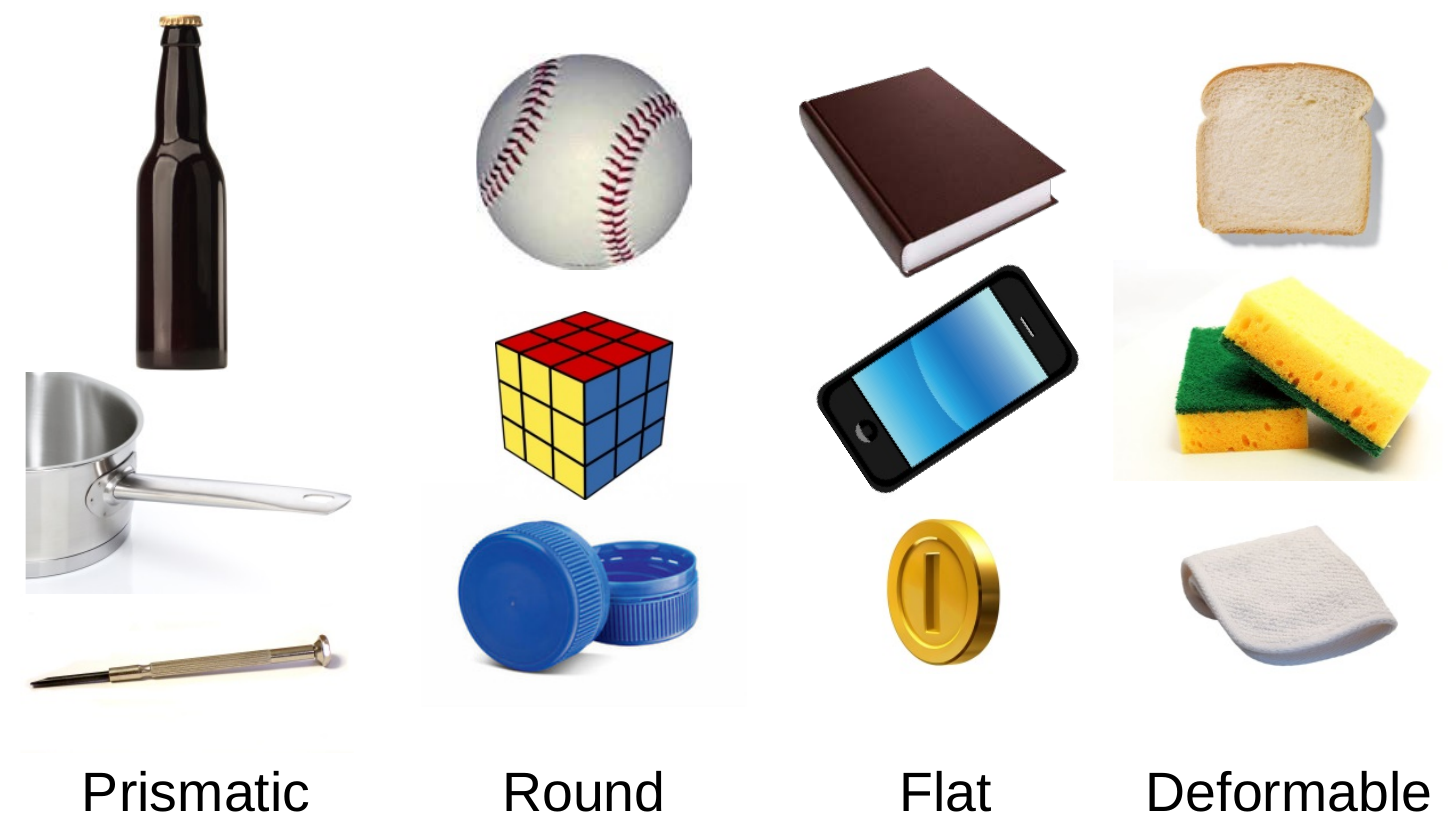}
    \caption{Example images of four different attributes: \textit{prismatic, round, flat,} and \textit{deformable}.}
    \label{fig_object_attribute}
\end{figure}

Attributes of the manipulated object are also important for understanding HOM activities since they reflect possible grasp types and actions that are affordable. Given an object with long and thin shape (like a pen), the hand is more likely to perform the action of \textit{writing} instead of \textit{drinking}.
While objects can be assessed by a wide range of attributes (shape, weight, surface smoothness, etc.), we only focus on attributes that are relevant to grasping and are also possible to be learned from image appearance. Figure~\ref{fig_object_attribute} shows some examples of the attributes studied in this work, three of which are related to object shape and the fourth is related to object rigidity. We identify three different shapes based on the criterion in Table~\ref{table_shape}. The fourth attribute of \textit{deformable} identifies the object that deforms under normal grasping forces. Examples are a sponge or a rag. In this work, we aim to characterize and classify the manipulated object based on the composition of these attributes.

\begin{table}[ht]
\caption{Classification criterion of three shape classes. Length of object along three object dimensions (major axes of the object) are denoted as $A$, $B$, and $C$, where $A\geq B\geq C$.}
\label{table_shape}
\small
\begin{center}
\begin{tabular}{|c|c|}
\hline
 Shape classes & Object dimensions \\
\hline
 Prismatic & $A>2B$ \\
 Round & $B\leq A<2B,\ C\leq A<2C$ \\
 Flat & $B>2C$ \\
\hline
\end{tabular}
\end{center}
\end{table}

\subsection{Training detectors}
\label{ssec_train_detector}
\begin{figure}[t]
    \centering
    \includegraphics[width=0.6\linewidth]{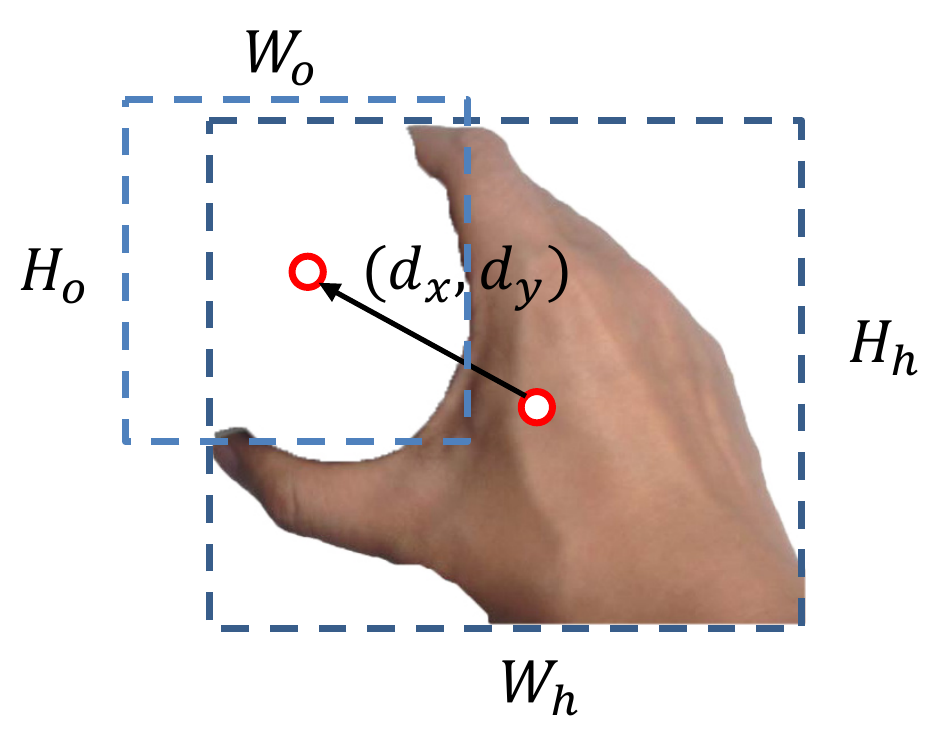}
    \caption{Illustration of relative location and scale of the hand and the grasped object.}
    \label{fig_target_regression}
\end{figure}

Object detection is a challenging task in computer vision, particularly unreliable when there are occlusions during manipulation. In this work, the grasped object is detected indirectly from current hand location and hand appearance. As illustrated in Figure~\ref{fig_target_regression}, relative location $(d_x,d_y)$ from the center of hand to the center of object is consistent to the hand orientation, and the object scale $(W_o,H_o)$ is related to the size of hand opening.
We formulate the hand and object detection as a two-stage problem. The first stage problem is hand segmentation which decides whether a pixel belongs to a hand, while the second stage problem is hand detection and object regression which detects real hand regions from hand segments and estimates object regions from hand regions. 

For hand segmentation, we train a Fully Convolutional Network (FCN) which takes as input an image and outputs a hand probability map. FCN is a transformed architecture of existing CNNs to output spatial maps instead of classification scores and is commonly used in semantic segmentation. The training data contains a pair of image and annotated ground-truth hand mask. The loss function for hand segmentation network is the sum of per-pixel two-class (background and hand) softmax loss.
Based on the obtained probability map, we can generate hand segments (blobs) by thresholding the probability map and get hand region proposals by drawing bounding boxes around the end of each hand blob. Hand region proposals are used as input of the second stage during model inference.
%and hand patches are then segmented with a bounding box. In detail, candidate hand regions are first selected by binarizing the probability map with a threshold. Regions under a certain area proportion are discarded and at most two regions are retained. Ellipse parameters (length of long/short axis, angle) are fitted to the hand region and the arm part is approximately removed by shortening the length of long axis to $1.5$ times of the length of short axis. Then the remaining region is cropped with a bounding box.

In the second stage, we train a Region-based Convolutional Neural Network (R-CNN) for hand and object detection (denoted as ``DetectionNet''). The DetectionNet takes a hand region as input and has two sibling output layers. The first sibling layer outputs classification scores over three categories: ``Background'', ``LeftHand'', and ``RightHand''. The reason why we set a category of ``Background'' for hand region proposals is to differentiate the skin-like non-hand background falsely segmented in the first stage. The second sibling layer outputs bounding-box regression offsets, $offset=(N_x,N_y,N_w,N_h)$, for the object grasped by the left hand and right hand respectively. The parameterization for $offset$ is given in the following equation set: 

\begin{equation}
\label{equation_target}
\begin{split}
N_x\ &=\ \frac{d_x}{W_h}, \ N_y\ =\ \frac{d_y}{H_h} \\
N_w\ &=\ \frac{W_o}{W_h}, \ N_h\ =\ \frac{H_o}{H_h}
\end{split}
\end{equation}
in which $N_x$ and $N_y$ specify scale-invariant transition from the center of hand to the center of object. $N_w$ and $N_h$ specify the scale ratio of object width and height with respect to the hand region.

Each training sample of DetectionNet is an image region labeled by a ground-truth hand label and a ground-truth bounding-box regression offset of the target object. We uses a multi-task loss as in \cite{girshick15fastrcnn}.
At testing, the output of classification layer is used as $\phi_{h}(P_h^s,I)$ in equation (\ref{equation_grasp}). The predicted offsets are used to calculate the standard object location as denoted by $P_{h\to o}^s$ in equation (\ref{equation_spatial}). 

\subsection{Training classifiers}
\label{ssec_train_classifier}
To this extent, we have trained a hand segmentation network and a detection network to localize the hands and the manipulated objects. We now move forward to introduce  
the procedure of training appearance-based classifiers to provide visual evidence for grasp types, object attributes and actions.

With the cropped images of hands, we train grasp CNN (denoted as ``GraspNet'') to recognize the grasp types. The network takes pairs of cropped hand regions and ground-truth grasp type labels as training data, and output a vector of classification scores for different grasp types. The output is used as $\phi_{g}(P_h^s)$ in equation (\ref{equation_grasp}).
Similar to GraspNet, we train object CNN (denoted as ``ObjectNet'') to recognize object attributes with the cropped images of objects. For training ObjectNet, pairs of cropped object regions and ground-truth object attribute labels are used as training data. The output is a vector of classification scores for different object attributes used as $\phi_{o}(P_o^s)$ in equation (\ref{equation_attribute}).

Unlike GraspNet and ObjectNet which use straightforward appearance cues from hand and object regions, the cues for training action classifier are more complex, involving both appearance and motion features. In this work, we train action CNN (denoted as ``ActionNet'') to recognize manipulation actions with two kinds of input. When action classification is conducted on single image, the whole image is used as input. When action classification is conducted on consecutive image sequence, optical flow images are used as input. The output of ActionNet is a vector of classification scores for different action classes used as $\phi_{a}(I)$ in equation (\ref{equation_action}).

\subsection{Learning model parameters}
\label{ssec_train_parameter}

Besides detectors and classifiers which are trained separately, model parameters are jointly learned from training data which weight different components of the model. At the parameter learning stage, annotations of hand and object bounding boxes, labels of grasp types, object attributes, and actions are used. 
We apply the DetectionNet and GraspNet to annotated hand bounding boxes to get hand detection scores and classification scores for the annotated grasp types. We apply the regression part of DetectionNet to hand bounding boxes together with the annotated object bounding boxes to get object detection scores, and apply ObjectNet to object bounding boxes to get classification scores for the annotated object attributes. We apply ActionNet to the whole image to get classification scores for the annotated action class.
Then, we use a maximum likelihood approach to estimate the optimal model parameters $\{\alpha,\beta,\gamma,\zeta,\eta,\lambda,\xi\}$ with which the model potential (\ref{equation_potential}) achieves highest score on the training data.

\subsection{Implementation details}
\label{ssec_implementation}
Here we give the details of network architecture and training for CNNs. Since a better architecture of CNN is out of the scope of this work, we choose the networks that are commonly used with the consideration of practical performance in different tasks.

We use SegNet (\cite{badrinarayanan2017segnet}) as the architecture for FCN-based hand segmentation. SegNet is an encoder-decoder architecture for semantic segmentation and we adopt it for its good balance on accuracy and efficiency. The network has input size of $K\times 3\times 360\times 480$ and output size of $K\times 2\times 360\times 480$ where $K$ is the batch size. We use a batch size of 8 and a fixed learning rate of $\gamma=0.1$ for training the network.  

We use CaffeNet (essentially AlexNet from \cite{krizhevsky2012imagenet}) as the architecture for hand/object detection. CaffeNet is a small-size network which contains five convolutional layers and three fully-connected layers. We choose this network due to the limited size of annotated hand and object bounding boxes. The same network is used for grasp type classification and object attribute classification. The network has input size of $K\times 3\times 227\times 227$. For hand/object detection, the output layer is replaced with two sibling layers as described earlier (Section \ref{ssec_train_detector}). For grasp type classification and object attribute classification, the output layer is replaced with a fully-connected layer and softmax over the number of categories. Due to the limited size of training samples, we adopt the finetuning approach and initialize CaffeNet using a pre-trained model from a large-scale dataset (ImageNet \cite{deng2009imagenet}). We use a batch size of 100 and a fixed learning rate of $\gamma=1e-4$ for finetuning.

We use VGG-16 (\cite{simonyan14c}) as the network architecture for action classification. VGG-16 is a deep CNN model composed by 16 weight layers and has become popular due to its high performance achieved on ImageNet classification task. The network has input size of $K\times C\times 224\times 224$, where $K$ is the batch size and $C$ is the number of channels. We set $C=3$ for the input of a raw image, and $C=20$ for the input of optical flow images. We finetune VGG-16 using pre-trained models from ImageNet and UCF101 (\cite{soomro2012ucf101}) for input of raw image and optical flow images respectively. We use a batch size of 25 and a fixed learning rate of $\gamma=5e-4$ for finetuning.

\section{Inference}
\label{sec_inference}

Given a new image, we obtain the results for hand and object detection, as well as the classification of grasp types, object attributes and actions by inference on equation (\ref{equation_potential}). We first initialize the model to get candidate hand and object locations and initial classification results (Section \ref{ssec_initialization}). Then we utilize an iterative method to update the inference results (Section \ref{ssec_iterative_inference}).

\subsection{Model initialization}
\label{ssec_initialization}

To initialize the model inference, we first use the trained segmentation network to get a set of candidate hand bounding boxes for each hand blob. Then we use the trained detection network to identify the hand side and remove falsely segmented background. Using the same network, we also get a set of candidate object bounding boxes for each valid hand bounding box. Finally, we initialize the classification results by applying the trained classifiers to the candidate hand and object bounding boxes and the current image respectively. Details of model initialization are given in the following steps.

\textbf{Initializing hand locations.} Hand blobs are first generated by thresholding the hand probability map obtained with the trained SegNet. Blobs beyond a certain range of area proportion are considered as false positive segments and discarded. Ellipse parameters (length of long/short axis, angle) are fitted to each hand blob and the forearm part is approximately removed by shortening the length of long axis to $1.5$ times of the length of short axis. The remaining region of each blob is cropped with a bounding box as a reference hand bounding box ($X_{ref},Y_{ref},W_{ref},H_{ref}$). For each reference hand bounding box, we further generate a set of candidate hand bounding boxes by sliding the reference bounding box in horizontal and vertical direction with a step of $\{-W_{ref}/8,0,W_{ref}/8\}$ and $\{-H_{ref}/8,0,H_{ref}/8\}$ respectively and also with a scale of $\{0.75,1.0,1.25\}$ for the width and height of each sliding window. We apply the trained R-CNN to all candidate hand bounding boxes to identify the hand side and remove the ones whose detection scores are below $0.8$.

\textbf{Initializing object locations.} For each remained hand bounding box, we first estimate the reference object bounding box via the regression part of DetectionNet. Then we generate a set of candidate object bounding boxes for each reference object bounding box following the same routine as in generating candidate hand bounding boxes. 

\textbf{Initializing classification.} We apply GraspNet to compute grasp type classification scores for each hand bounding box, apply ObjectNet to compute object attribution classification scores for each object bounding box, and apply ActionNet to compute action classification scores for current image. The labels of grasp type and object attribute are initialized with the class that has the highest score on the reference hand and object bounding boxes respectively. The action label is initialized with the class that has the highest action classification score.

\subsection{Iterative inference}
\label{ssec_iterative_inference}

After model initialization, we iteratively update the inference results by performing the following steps.

\textbf{Updating hand detection and grasp type classification.} Based on the current results of action and object attribute labels, we update hand locations and labels of grasp type. The score of assigning grasp type $g_i$ to the hand bounding box $P_h^l$ of the left-side hand given action label $a_k$, grasp type label $g_j$ at the other side and object attribute label $o_m,o_n$ at both sides is represented as:
\begin{equation}
\label{equation_infer_grasp}
\begin{split}
f(i,P_h^l)=&\alpha_{k,i,j,m,n}+\beta_{i,m}+\zeta_l^T\cdot \phi_{h}(P_h^l,I)\\
&+\eta_{i}^T\cdot \phi_{g}(P_h^l).    
\end{split}
\end{equation}
The hand location and grasp type label are updated by selecting $(i^*,P_h^{l*}) = \argmax f(i,P_h^l)$. The same inference is conducted for the right-side hand.

\textbf{Updating object detection and object attribute classification.} We first obtain the reference object bounding box $P_{h\to o}^l$ and a set of candidate object bounding boxes with the current bounding box of the left-side hand. Then, based on the current action label $a_k$, the object attribute label $o_n$ at the other side and the grasp type label $g_i,g_j$ of two sides, we assign object attribute $o_m$ to the object bounding box $P_o^l$ by optimizing the following score function:
\begin{equation}
\label{equation_infer_object}
\begin{split}
f(m,P_o^l)=&\alpha_{k,i,j,m,n}+\beta_{i,m}+\gamma_l\cdot p(P_o^l|P_{h\to o}^l)\\
&+\lambda_{j}^T\cdot\phi_{o}(P_o^l).    
\end{split}
\end{equation}
The same inference is conducted for the right-side object.

\textbf{Updating action classification.} Based on the current classification results of grasp type and object attribute, we update the action label by optimizing $\Psi_{FC}(G,O,A)+\Psi_{A}(A,I)$ with the enumeration of all possible action classes.

\section{Evaluation}
\label{s_experiment}

In this section, we evaluate our method on two public datasets. Datasets are first introduced in Section~\ref{ss_dataset}. Visualization of the learned model is shown in Section~\ref{ss_context}. Performance of the three recognition tasks are presented in Section~\ref{ss_hand_object} (grasp type and object attribute) and Section~\ref{ss_action} (action). Finally, generalizability study is given in Section~\ref{ss_general}.

\subsection{Dataset}
\label{ss_dataset}

\begin{table*}[ht]
\caption{Annotations of grasp types, object attributes and actions in GTEA Dataset. Action labels are provided in the dataset. Grasp types and object attributes are newly annotated in this work.}
\label{table_annotation}
\begin{center}
\begin{tabular}{|ccc|ccc|ccc|}
\hline
 \multicolumn{3}{|p{5cm}|}{Grasp type} & \multicolumn{3}{|p{5cm}|}{Object attribute} & \multicolumn{3}{|p{4cm}|}{Action}\\
 label & description & ratio & label & description & ratio & label & description & ratio\\
\hline
 g01 & power wrap & 24.0\% & o01 & rigid-prismatic & 25.4\% & a01 & close & 10.0\%\\
 g02 & precision wrap & 8.8\% & o02 & deformable-prismatic & 6.5\% & a02 & fold & 1.5\%\\
 g03 & parallel extension & 18.7\% & o03 & rigid-flat & 8.4\% & a03 & open & 23.9\%\\
 g04 & platform & 4.9\% & o04 & deformable-flat & 20.9\% & a04 & pour & 13.1\%\\
 g05 & precision sphere & 6.1\% & o05 & rigid-round & 12.1\% & a05 & put & 14.7\%\\
 g06 & power sphere & 5.8\% & o06 & deformable-round & 1.5\% & a06 & scoop & 13.1\%\\
 g07 & index finger extension & 7.5\% & o07 & rigid-prismatic-flat & 16.0\% & a07 & shake & 0.4\%\\
 g08 & writing tripod & 5.1\% & o08 & rigid-flat-small & 4.4\% & a08 & spread & 9.7\%\\
 g09 & lateral pinch & 4.8\% & o09 & deformable-flat-small & 4.9\% & a09 & stir & 3.9\%\\
 g10 & thumb-4 finger & 2.0\% & & & & a10 & take & 9.7\%\\
 g11 & thumb-2 finger & 5.5\% & & & & & & \\
 g12 & tip pinch & 4.6\% & & & & & & \\
 g13 & inferior pincer & 2.2\% & & & & & & \\
\hline
\end{tabular}
\end{center}
\end{table*}

We mainly evaluate our approach on the public GTEA Dataset (\cite{fathi2011understanding}) recorded with a head-mounted wearable camera. This dataset consists of 7 different cooking activities performed by 4 different subjects. Action annotations are provided with each containing a verb (action) and a set of objects with the beginning and ending frame. There are 10 action categories and 525 instances (video segments) in the original labels. To train and test the proposed model, we further annotated 1583 images with the following information: bounding boxes and grasp type labels of hands, bounding boxes and object attribute labels of the grasped objects. In Table~\ref{table_annotation}, we summarize the name, proportion of each class for grasp types, object attributes and actions. Experiments of the three recognition tasks are evaluated using leave-one-subject-out cross-validation. 

We also conduct experiments on GTEA Gaze Dataset (\cite{fathi2012learning}) to study the generalizabiity of our method. This dataset consists of 17 sequences of cooking activities performed by 14 different subjects, also recorded with a wearable camera. The action verb and object categories with beginning and ending frame are annotated. Main difference of the two datasets is that they are recorded on different environments with different wearable devices, thereby causing appearance variation on similar activities. With similar annotation criterion, we further annotated 732 images for the experiments. We adopt the train/test splits for training and testing following previous work.

\subsection{Learned context}
\label{ss_context}

\begin{figure*}[t]
    \centering
    \includegraphics[width=\linewidth]{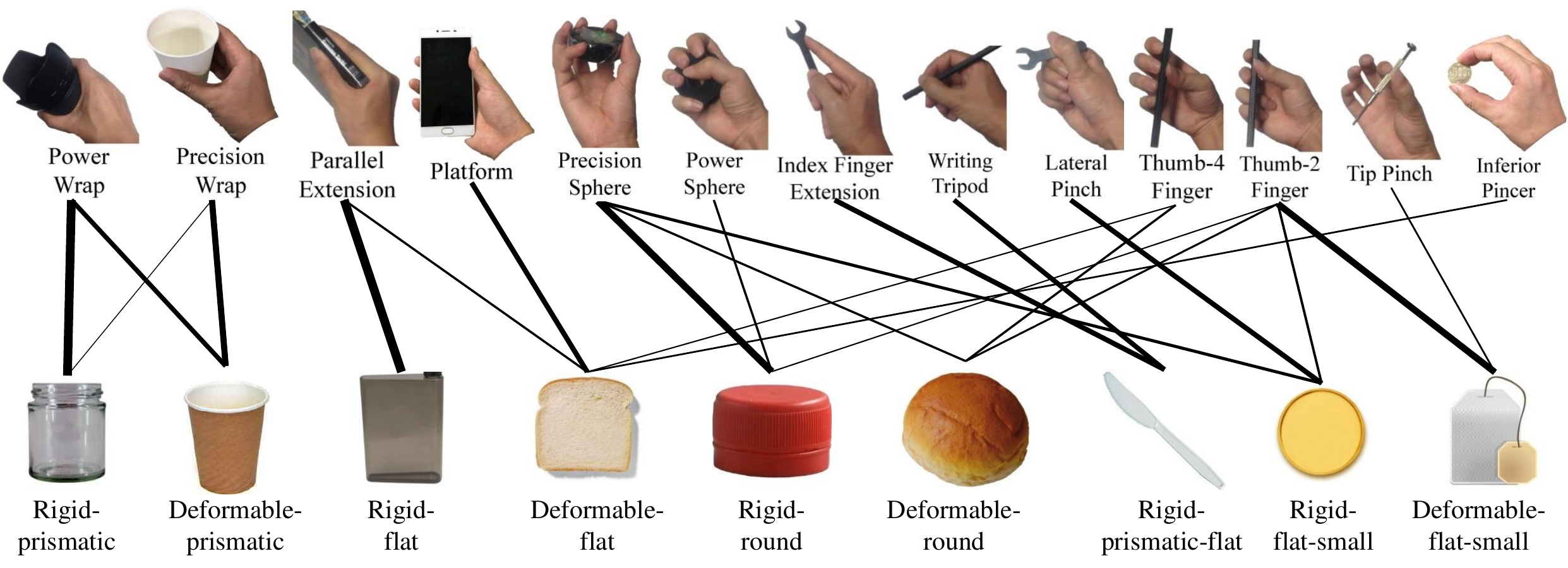}
    \caption{Visualization of the physical context learned on GTEA Dataset. The line strength indicates the strength of relationship between two classes.}
    \label{fig_context_go}
\end{figure*}

Based on visual evidence ($\Psi_G, \Psi_O, \Psi_A$ in (\ref{equation_potential})) of 13 grasp types, 9 object attributes and 10 manipulation actions, our model learns contextual relationships ($\Psi_{FC},\Psi_{PC},\Psi_{SC}$ in (\ref{equation_potential})) between these components. In this section, we show the visualization of semantic context ($\Psi_{FC}$ and $\Psi_{PC}$) learned from the GTEA Dataset.

\begin{figure*}[t]
    \centering
    \includegraphics[width=\linewidth]{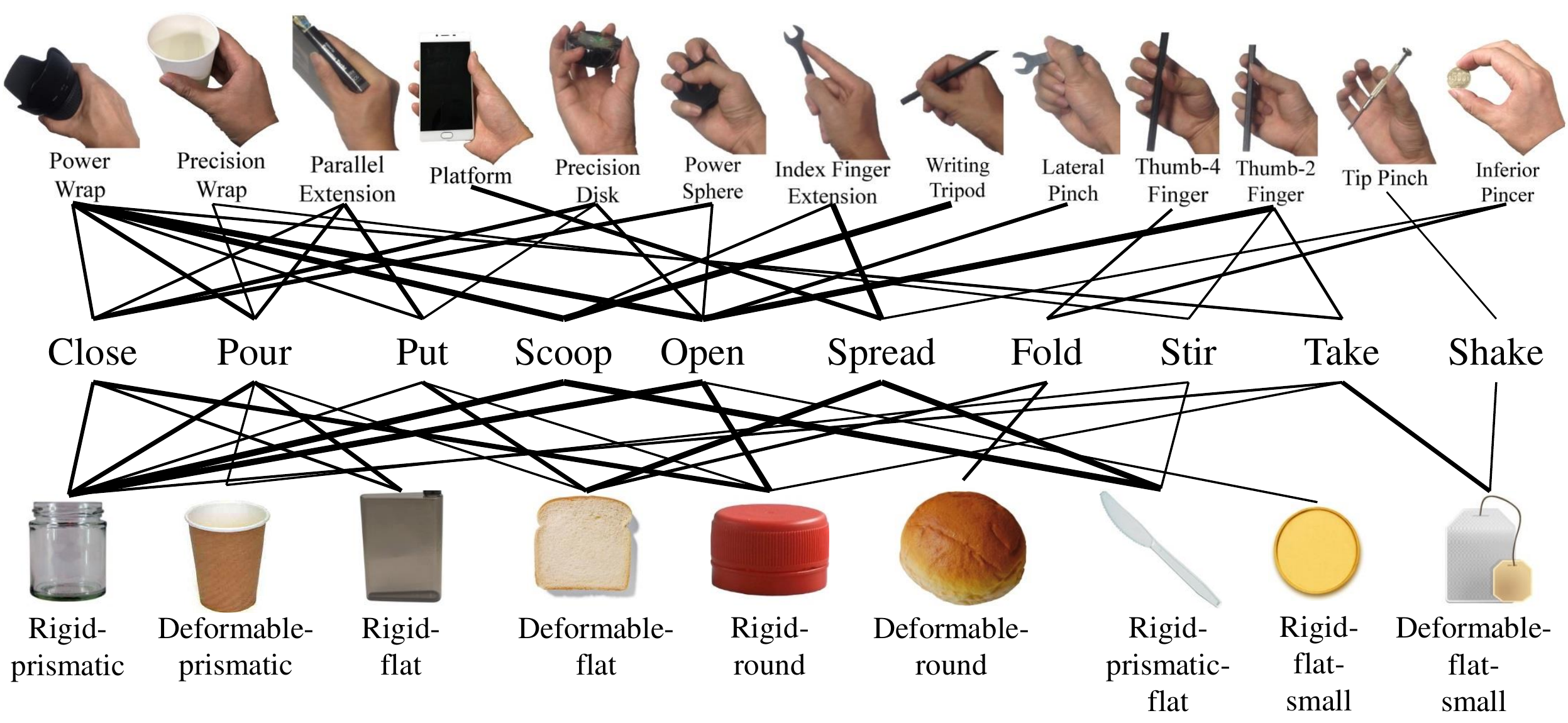}
    \caption{Visualization of the functional context learned on GTEA Dataset. The line strength indicates the strength of relationship between two classes.}
    \label{fig_context_ago}
\end{figure*}

\begin{figure*}[t]
    \centering
    \includegraphics[width=\linewidth]{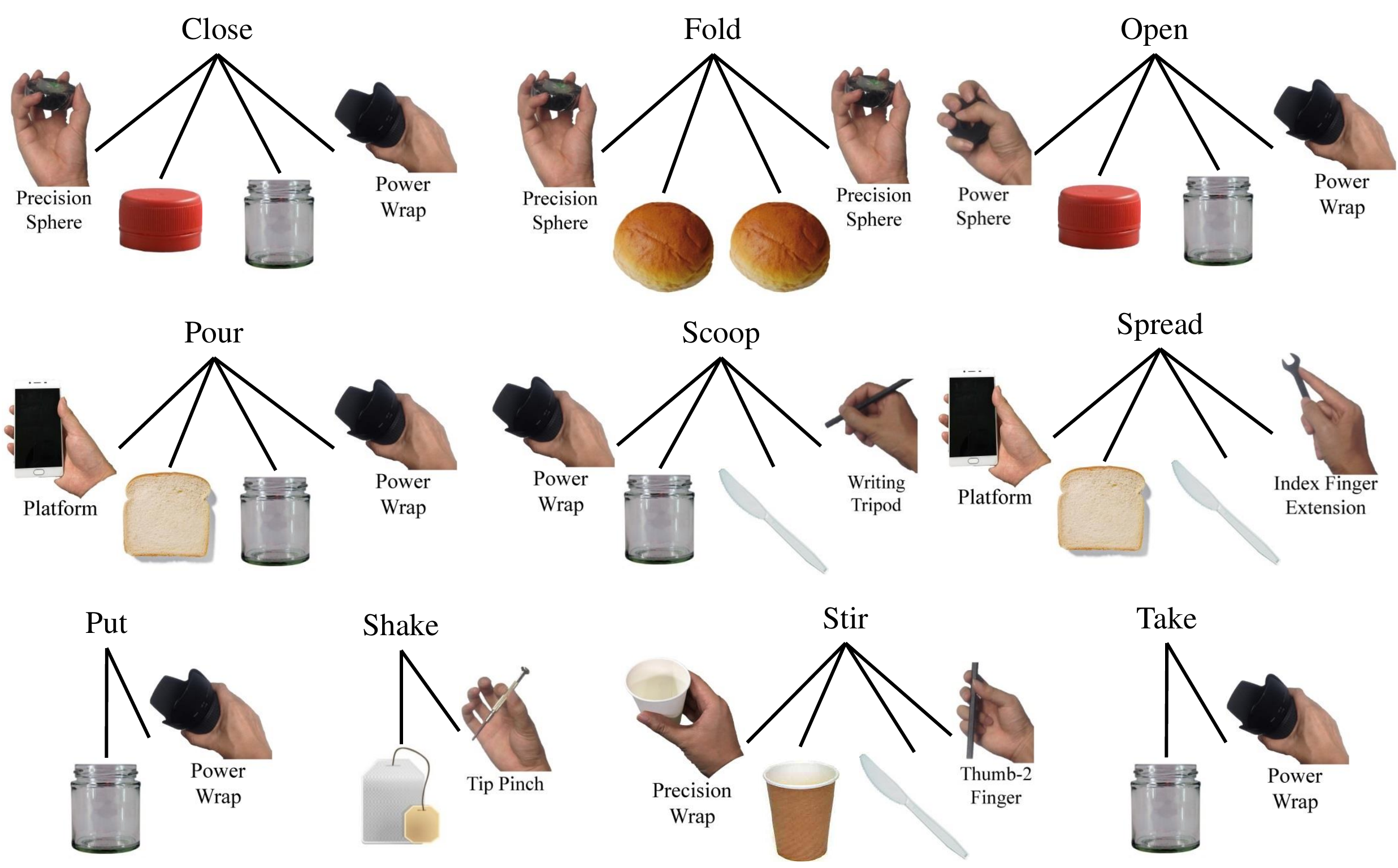}
    \caption{Visualization of the most probable combinations of grasp types and object attributes conditioned on each action class.}
    \label{fig_context_demo}
\end{figure*}

Visualization of the learned physical context between grasp types and object attributes ($\Psi_{PC}$) is shown shown in Figure~\ref{fig_context_go}. The line strength in the figure demonstrates the strength of the context, which is estimated from $\beta$ in (\ref{equation_go}). It can be seen that grasp types and attributes of the grasped objects are strongly correlated. For example, there is a strong connection between the \textit{rigid-prismatic} object (such as a glass bottle) and the grasp type of \textit{power wrap} which is the main hand pose affordable to hold the object stably in hand. Other strong connections can be found between the \textit{rigid-flat} object and \textit{parallel extension}, the \textit{rigid-round} object and \textit{precision sphere}, etc. The correlation between hands and objects helps us to better recognize grasp types and object attributes together. For example, it is more likely that the \textit{rigid-round} object is grasped by \textit{precision sphere} than by \textit{thumb-2 finger}. From Figure~\ref{fig_context_go}, we also observe dexterous hand-object interactions that the same object can be held by multiple grasp types and the same grasp type affords to manipulate more than one types of objects. Nevertheless, the dexterous aspect of hand-object interactions provides important information for understanding manipulation activities. For example, the connections between the \textit{rigid-prismatic-flat} object and two different grasp types are related to two different actions: \textit{scoop} and \textit{spread} as occurred in the dataset.  

Visualization of the learned functional context between actions, grasp types and object attributes ($\Psi_{FC}$) is shown in Figure~\ref{fig_context_ago}. Specifically, we visualize the strength of action-hand and the action-object relationships separately, which are estimated by marginalizing $\alpha$ in (\ref{equation_ago}) with respect to object attributes and grasp types respectively. Figure~\ref{fig_context_ago} demonstrates that our model learns meaningful hand-action-object interactions in manipulation activities. One important observation is that grasp type together with object attribute provide complementary information which is useful in characterizing different actions. Taking \textit{deformable-flat-small} object (such as a tea pack) for example, it is grasped with \textit{thumb-2 finger} during the action of \textit{take}, while grasped with \textit{tip pinch} during the action of \textit{shake}. Furthermore, we observe that some actions involve complex hand-object interactions and information from both hands are needed to distinguish between them. Taking \textit{close} and \textit{pour} for example, although both actions involve a hand holding a bottle-like object on one side, on the other side different objects (such as \textit{rigid-round} bottle cap and \textit{deformable-prismatic} cup) are held by different grasp types (such as \textit{precision sphere} and \textit{precision wrap}) respectively. Nevertheless, these complex interactions are learned in our proposed functional context. 
Figure~\ref{fig_context_demo} illustrates the most likely combinations of grasp types and object attributes for each action class. It can be seen that 7 out of 10 action classes are most likely characterized by bimanual operation. 

In the following sections, we will show how the learned context helps improve the recognition performance of grasp types, object attributes and actions.

\begin{figure*}[t]
    \centering
    \includegraphics[width=0.9\linewidth]{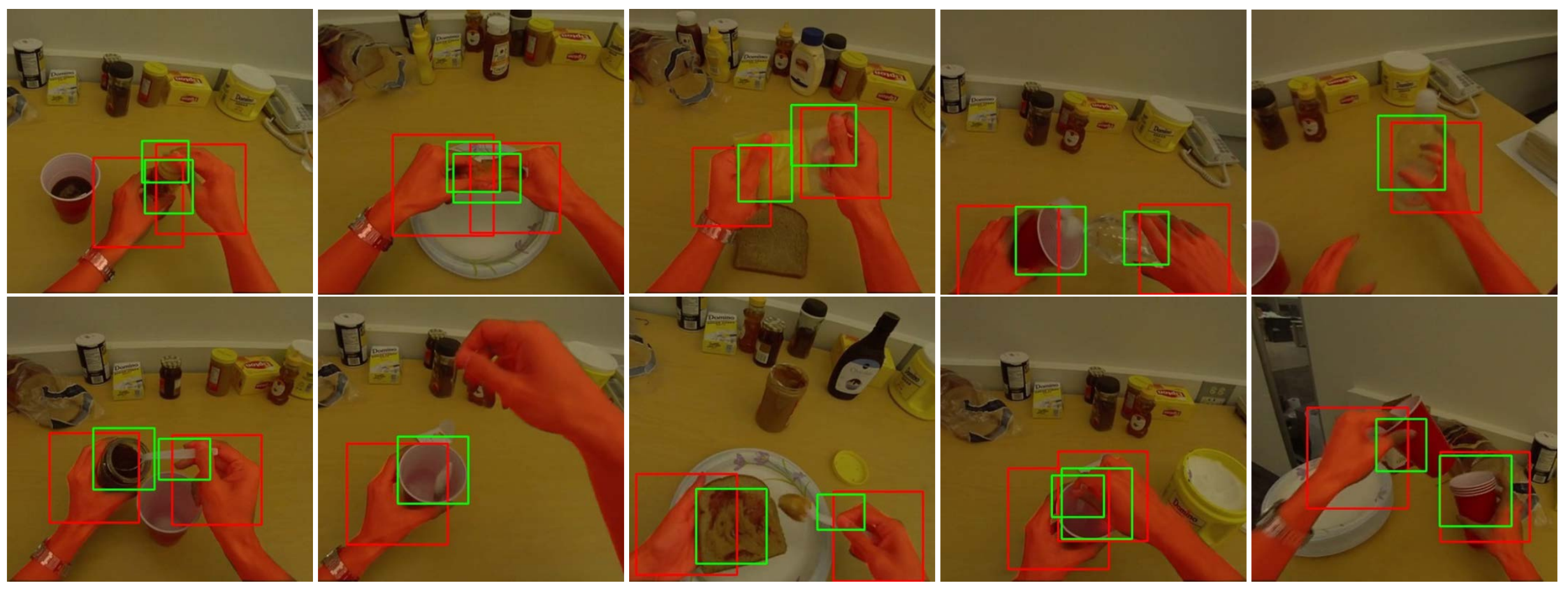}
    \caption{Hand and object detection on GTEA Dataset. Hand probability map (\textcolor{red}{red}) is overlapping with the raw image. Hand bounding box (\textcolor{red}{red}) and object bounding box (\textcolor{green}{green}) are draw on top of the image. Examples are selected from 10 different actions (Top 5: close, fold, open, pour, put; Bottom 5: scoop, shape, spread, stir, take).}
    \label{fig_detect_result}
\end{figure*}

\subsection{Recognition of grasp type and object attribute}
\label{ss_hand_object}

\textbf{Hand and object detection.} As described in Section~\ref{ssec_train_detector}, hand and object detection is formulated as a two-stage problem. In the first stage, we train a hand segmentation network to generate hand probability map for each image which is then used for computing hand region proposals. We use raw images and hand masks provided with GTEA Dataset as training data for hand segmentation network. The input image dimension is down-sampled to the input size of the network and the output probability map is up-sampled to the original size. Hand region proposals are generated by thresholding the probability map and drawing bounding boxes around the ends of each hand blob. In the second stage, we train a detection network to filter candidate hand bounding boxes and regress the bounding box offsets for the grasped objects. The generated candidate hand and object bounding boxes are used as input of the model inference and the final hand/object locations are estimated with corresponding grasp types/object attributes.

Qualitative results of hand and object detection in different actions are shown in Figure~\ref{fig_detect_result}. The detected hand and object bounding boxes match well with the real hand and object locations, even though the background is cluttered and the grasped objects are partially occluded. The results demonstrate the effectiveness of hand segmentation network and the feasibility of detecting the grasped objects simply from hand appearance.

\textbf{Grasp type recognition.} Here we evaluate the recognition performance of $13$ grasp types on GTEA Dataset. We use GraspNet as the baseline in which the grasp type is assigned to the category with which highest score obtains in the output softmax layer. Note that the GraspNet is also used for initializing the grasp classification scores in equation (\ref{equation_grasp}) of our model. We also compare our method with another baseline (denoted as ``GraspNet+object context'') in which the attribute information from the grasped object serves as object context to improve the recognition performance. This method was first used in the work of \cite{cai2016understanding}. 

Table \ref{table_grasp} shows grasp type recognition results of different methods. It can be observed that our method achieves the best performance. Compared with other baselines without or with limited context, our method explores the semantic relationship with the grasped object and the performed action, which helps to differentiate grasp types that are visually ambiguous. Taking g04 (\textit{platform}) for example, the GraspNet achieves accuracy of $75.0\%$, including context from object improves the performance to $78.2\%$, and our method further enhances the performance to $87.8\%$. Two reasons might explain the performance improvement. On the one hand, object attributes provide physical constraints on a small set of grasp types which are affordable to hold the object. On the other hand, our method also takes functional constraints by recognizing different manipulation actions (\eg, recognition of \textit{spread}), which further helps to differentiate visually similar grasp types. 

\begin{table}[ht]
\caption{Grasp type recognition results on GTEA Dataset. Classification accuracy is used as evaluation metric.}
\label{table_grasp}
\begin{center}
\begin{tabular}{|c|c|c|c|}
\hline
 \multirowcell{2}{Method} & \multirowcell{2}{GraspNet} & \multirowcell{2}{GraspNet+\\object context} & \multirowcell{2}{Our\\method}\\
\hline
\hline
 g01 & 65.5\% & 63.2\% & \textbf{68.3}\%\\
 g02 & 72.1\% & \textbf{74.5}\% & \textbf{74.5}\%\\
 g03 & 63.8\% & 71.4\% & \textbf{72.2}\%\\
 g04 & 75.0\% & 78.2\% & \textbf{87.8}\%\\ %010-030
 g05 & 37.1\% & 46.5\% & \textbf{48.0}\%\\
 g06 & 40.0\% & 42.8\% & \textbf{46.1}\%\\ %010-030
 g07 & 53.4\% & 54.5\% & \textbf{57.1}\%\\
 g08 & 68.8\% & 72.4\% & \textbf{73.6}\%\\
 g09 & 23.5\% & 23.5\% & \textbf{27.5}\%\\
 g10 & 44.4\% & \textbf{52.9}\% & 50.0\%\\
 g11 & 34.7\% & 35.2\% & \textbf{37.5}\%\\
 g12 & 37.5\% & 37.5\% & \textbf{48.0}\%\\ %010-030
 g13 & 21.0\% & \textbf{26.6}\% & 21.0\%\\
\hline
\hline
 Overall & 51.4\% & 55.5\% & \textbf{57.5}\%\\
\hline
\end{tabular}
\end{center}
\end{table}

\textbf{Object attribute recognition.} Similar to grasp type recognition, we compare our method with two baselines without (``ObjectNet'') or with limited context (``ObjectNet+grasp context''), and show how object attribute recognition is improved by contextual information from the grasp types and actions. As shown in Table \ref{table_object}, overall accuracy for $9$ different object attributes is $58.9\%$ from ObjectNet. Grasp context improves the accuracy by $2.9\%$, and our method further improves the performance by $2.1\%$. It is interesting to see that the performance improvement for o03 ($9.1\%$) is consistent with the that for g03 ($8.4\%$). Actually, g03 (\textit{parallel extension}) and o03 (\textit{rigid flat}) are strongly correlated in HOM activities and the mutual context learned between the two helps to enhance the recognition performance in both tasks. 

\begin{table}[ht]
\caption{Object attribute recognition results on GTEA Dataset. Classification accuracy is used as evaluation metric.}
\label{table_object}
\begin{center}
\begin{tabular}{|c|c|c|c|}
\hline
 \multirowcell{2}{Method} & \multirowcell{2}{ObjectNet} & \multirowcell{2}{ObjectNet+\\grasp context} & \multirowcell{2}{Our\\method}\\
\hline
\hline
 o01 & 65.7\% & 64.4\% & \textbf{69.2}\%\\
 o02 & 83.6\% & 85.1\% & \textbf{87.2}\%\\
 o03 & 70.9\% & 78.6\% & \textbf{80.0}\%\\
 o04 & 78.4\% & 85.1\% & \textbf{88.8}\%\\
 o05 & 43.8\% & 41.6\% & \textbf{45.9}\%\\
 o06 & 37.5\% & \textbf{42.8}\% & \textbf{42.8}\%\\ %010-030
 o07 & 63.9\% & 68.2\% & \textbf{69.0}\%\\
 o08 & 25.0\% & \textbf{30.7}\% & 28.5\%\\
 o09 & 14.8\% & 14.8\% & \textbf{22.2}\%\\
\hline
\hline
 Overall & 58.9\% & 61.8\% & \textbf{63.9}\%\\
\hline
\end{tabular}
\end{center}
\end{table}

\begin{figure}[t]
    \centering
    \includegraphics[width=\linewidth]{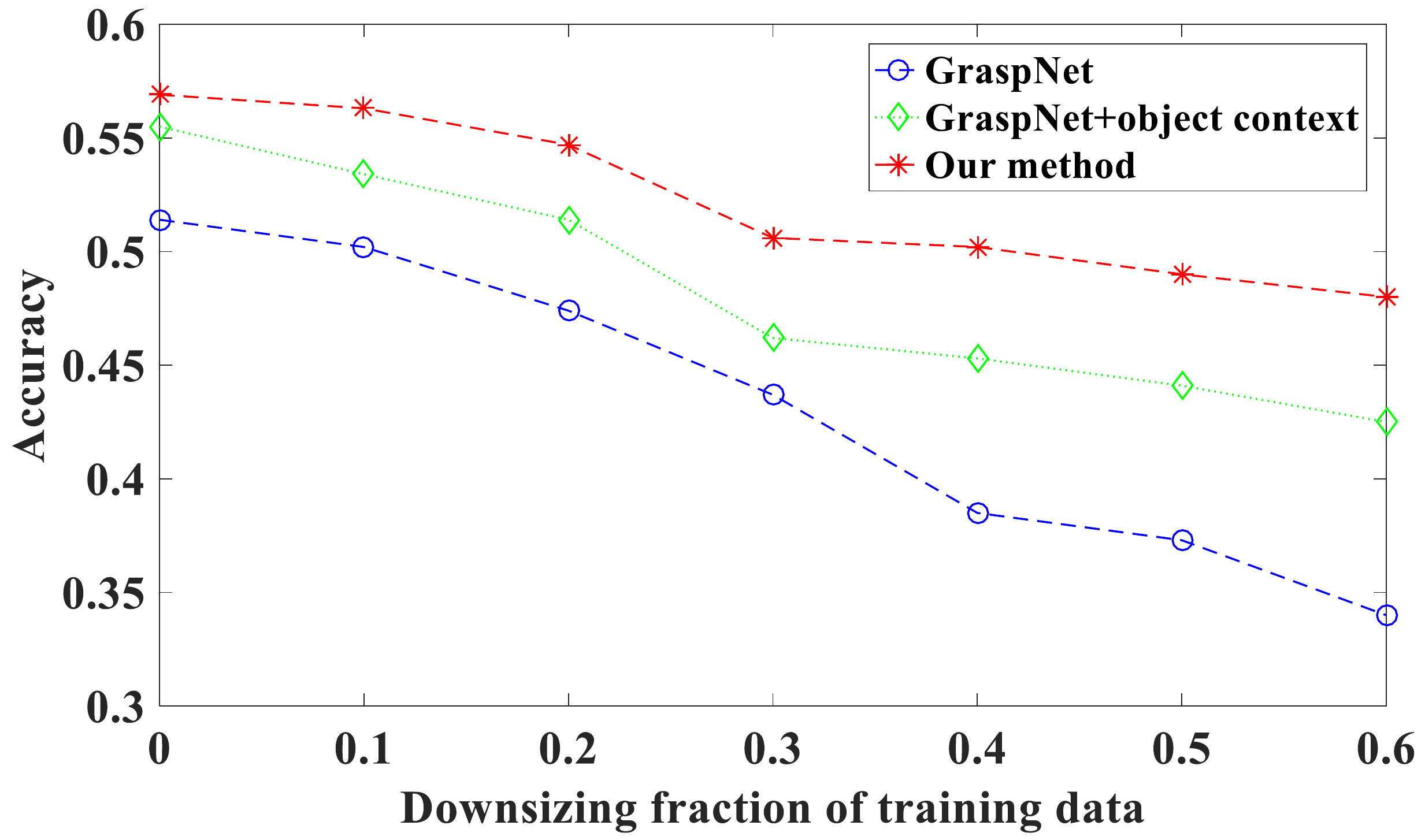}
    \caption{Performance variation of grasp type recognition at different sizes of training data. The horizontal axis shows the fraction of training data that are removed. Downsizing fraction of 0.1 means $10\%$ of training data are removed.}
    \label{fig_ablation_grasp}
\end{figure}

\begin{figure}[t]
    \centering
    \includegraphics[width=\linewidth]{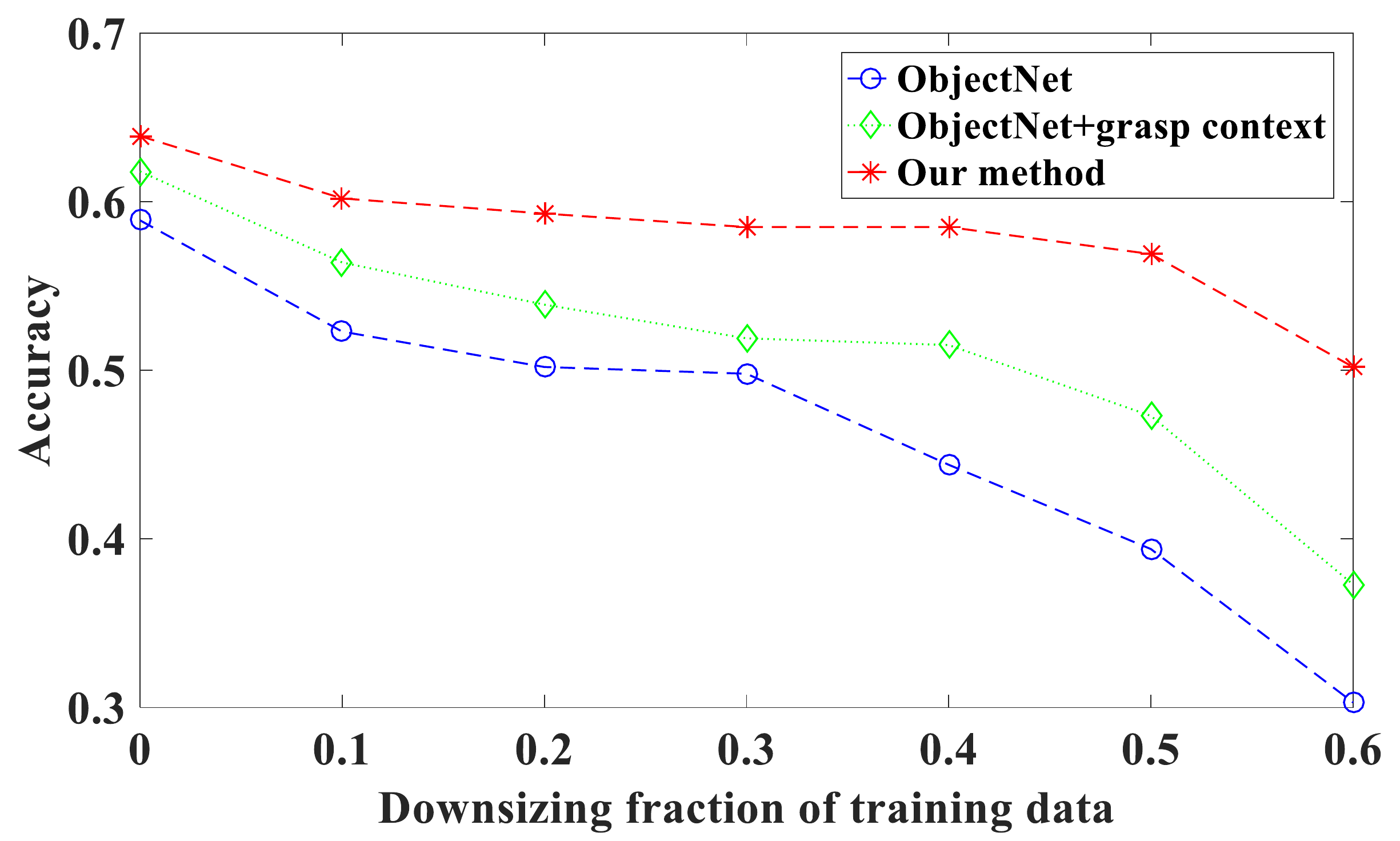}
    \caption{Performance variation of object attribute recognition at different sizes of training data.}
    \label{fig_ablation_object}
\end{figure}

\textbf{Robustness study.} We have shown in Table~\ref{table_grasp} and Table~\ref{table_object} that the recognition performance of both grasp types and object attributes have been improved by our method.
%Although the scale of improvement doesn't seem significant (around $5\%$), we think the reason might be that the CNN-based classifiers have already learned well the class difference from training data. 
To get a better understanding about the effectiveness of context, we further evaluate the recognition performance in difficult situations where annotated labels are insufficient to train reliable classifiers. Specifically, we gradually downsize the training data of grasp types and object attributes respectively and evaluate the performance of different methods. Figure \ref{fig_ablation_grasp} demonstrates the performance variation of grasp type recognition. As we increase the downsizing fraction, performance of all methods decreases. However, the methods using context (GraspNet+object context, and our method) degrade much less slowly than CNN-based method (GraspNet), and the performance gap between GraspNet and our method increases nearly to $15\%$ when $60\%$ of training data are removed. Similar performance variation for object attribute recognition can be observed in Figure \ref{fig_ablation_object}. The results demonstrate the importance of contextual relationships in multi-task recognition when sufficient annotations are not available for all tasks.

\subsection{Action recognition}
\label{ss_action}

In this section, we evaluate the action recognition performance of three different methods. The first method uses action CNN (ActionNet) to recognize actions directly from visual observation, often in terms of image appearance or optical flow-based motion. In \cite{ma2016going}, a two-stream CNN model incorporating information from both appearance and motion has achieved state-of-the-art performance in egocentric action recognition. Note that a better CNN architecture is not the focus of this work, and we are more interested in how contextual relationships could improve action recognition performance of different CNN models. Therefore, we train ActionNet based on single image and optical flow images separately (denoted as ``ActionNet-rgb'' and ``ActionNet-flow'' respectively). The second method models manipulation action only by its functional components of hands and objects, more specifically, of hand grasp types and object attributes. Instead of visual observation as used in the first method, only contextual information from grasp types and object attributes are used in the second method (denoted as ``Context only''). Finally, in our method, both visual observation and contextual information are utilized in the proposed model. We denote our methods as ``Ours-rgb'' and ``Ours-flow'', in which ActionNet-rgb and ActionNet-flow are used to provide visual evidence for actions respectively.

Table \ref{table_action} shows action recognition results on different action classes and the overall accuracy. ActionNet-rgb is used as baseline to show how performance could be improved for image-based recognition. Our method achieves best performance and outperforms ActionNet-rgb by over $10\%$, demonstrating the effectiveness of our model in image-based action recognition. It is also important to note that the second method with only contextual information outperforms ActionNet-rgb by $6.9\%$. Since the contextual information comes from visual recognition of hand and object regions, the result actually indicates that it is more efficient to model actions from appearance of interaction than from the whole image.

\begin{table}[h]
\caption{Image-based action recognition results on GTEA Dataset. Classification accuracy is used as evaluation metric.}
\label{table_action}
\begin{center}
\begin{tabular}{|c|c|c|c|}
\hline
 Method & ActionNet-rgb & Context only & Ours-rgb\\
\hline
\hline
 a01 & 29.0\% & 28.0\% & \textbf{37.7}\%\\
 a02 & 40.0\% & \textbf{100.0}\% & 66.6\%\\
 a03 & 56.6\% & 55.2\% & \textbf{64.3}\%\\
 a04 & 46.6\% & \textbf{78.6}\% & 71.8\%\\
 a05 & 67.9\% & 61.1\% & \textbf{69.2}\%\\
 a06 & 31.1\% & \textbf{76.7}\% & 56.0\%\\
 a07 & 23.9\% & 28.2\% & \textbf{29.0}\%\\
 a08 & 82.3\% & 64.2\% & \textbf{88.4}\%\\
 a09 & 42.8\% & \textbf{58.3}\% & 50.0\%\\
 a10 & 55.1\% & 51.0\% & \textbf{62.7}\%\\
\hline
\hline
 Overall & 53.3\% & 60.2\% & \textbf{63.7}\%\\
\hline
\end{tabular}
\end{center}
\end{table}

\begin{figure}[t]
    \centering
    \includegraphics[width=0.95\linewidth]{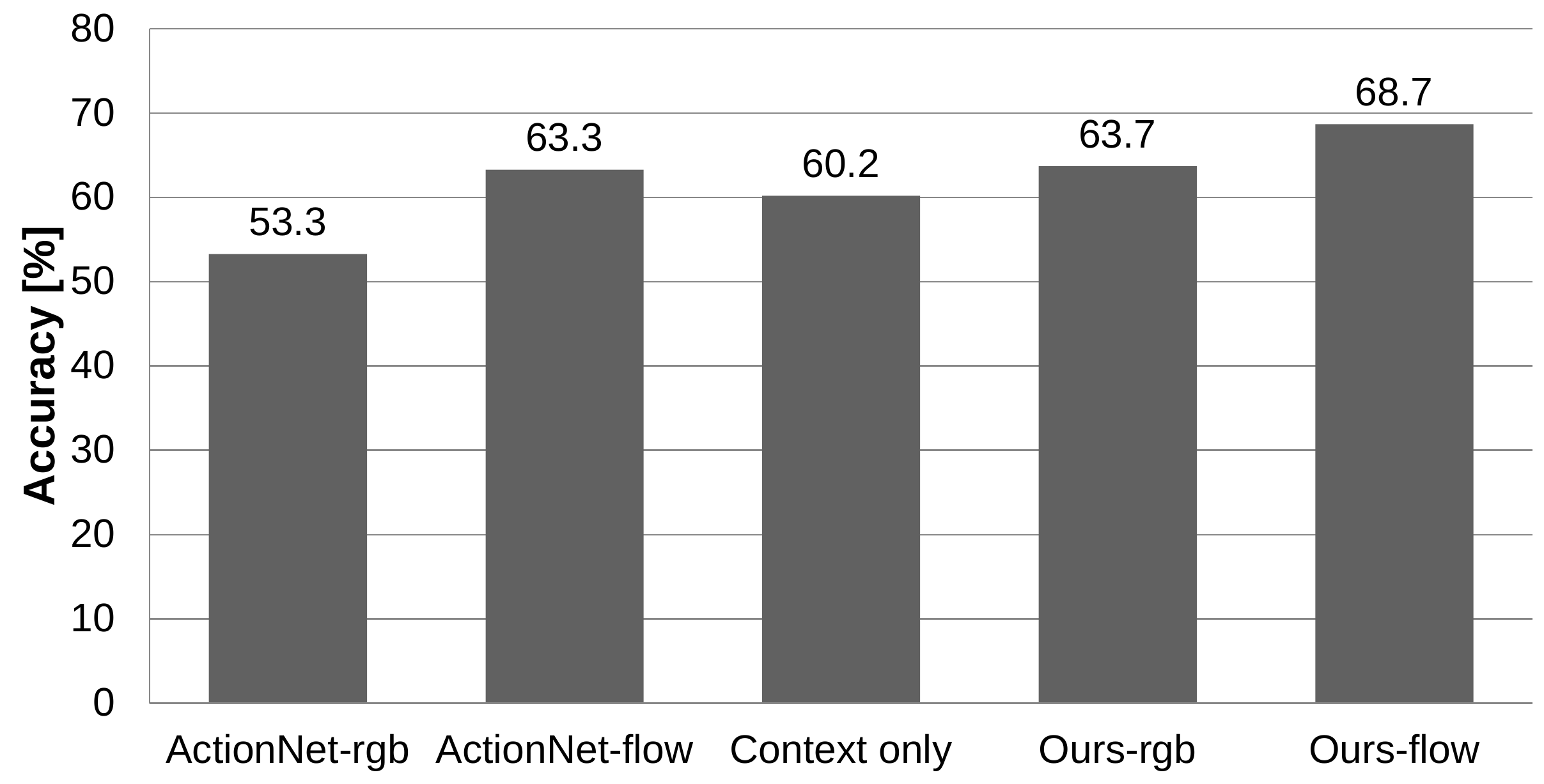}
    \caption{Action classification results of different methods on the GTEA dataset. Classification accuracy is used to evaluate the performance.}
    \label{fig_actionresult_GTEA}
\end{figure}

Figure \ref{fig_actionresult_GTEA} summarizes the overall performance of different methods. ActionNet-flow outperforms ActionNet-rgb by $10\%$, showing that motion information from multiple consecutive images has obvious advantage of discriminative power over appearance information from a single image. However, the performance of ``Ours-rgb'' is even slightly better than ActionNet-flow, indicating the ability of context to boost the image-based action recognition to the extent that can only be achieved from video. The best performance of $68.7\%$ accuracy is achieved from ``Ours-flow'', outperforming ActionNet-flow by $5.4\%$.

\subsection{Generalizability study}
\label{ss_general}
By far we have shown that our method with the learned contextual relationships has achieved significant performance improvement on GTEA Dataset. However, generalizability is always an issue that how the research findings and conclusions from a study on a sample population could generalize to a larger population. In this section, we study the generalizability of our method by evaluating its performance across two different datasets (GTEA Dataset and GTEA Gaze Dataset). 

\begin{figure}[t]
    \centering
    \includegraphics[width=0.95\linewidth]{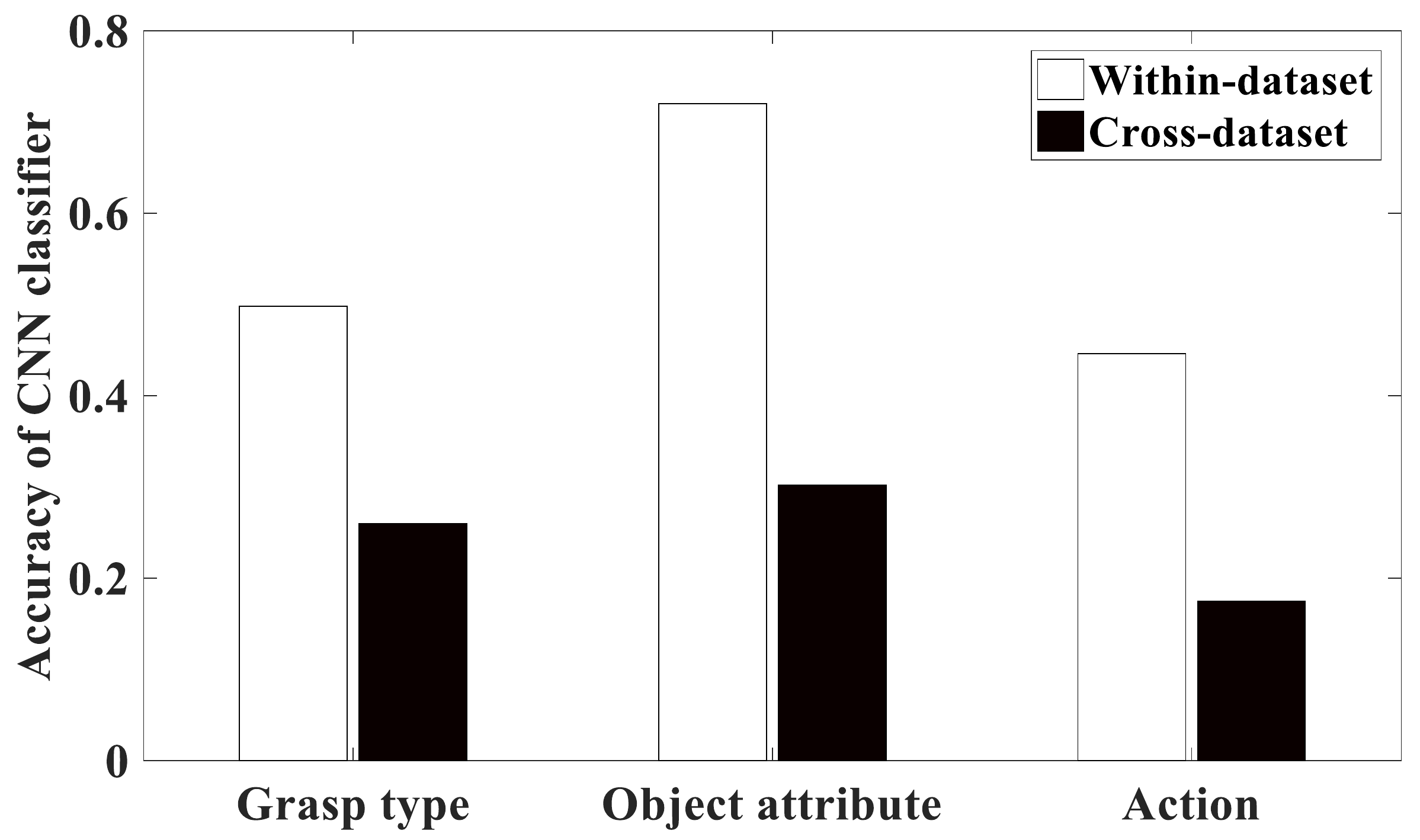}
    \caption{Generalizability evaluation of CNN-based classifiers by classification accuracy of grasp types, object attributes and actions. For ``within-dataset'', CNN classifiers are trained (or finetuned) and tested on the training/testing division of GTEA Gaze Dataset. For ``cross-dataset'', CNN classifiers are trained on GTEA Dataset and tested on GTEA Gaze Dataset.}
    \label{fig_cross_cnn}
\end{figure}

We first evaluate the cross-dataset performance of CNN-based classifiers. Although CNNs are not the main novelty of our method, it is important to know their generability across different datasets, as they are widely used in various recognition problems and also provide visual evidence in our method. Figure \ref{fig_cross_cnn} shows cross-dataset performance of CNN classifiers by classification accuracy of grasp types, object attributes and actions respectively. Large performance drop could be seen for all three classification tasks. The results indicate that CNN classifiers are sensitive to appearance variation (kind of overfitting to the training data), and can not be directly applied to other datasets without sufficiently large training data or finetuning on the target data space.

\begin{figure}[t]
    \centering
    \includegraphics[width=0.95\linewidth]{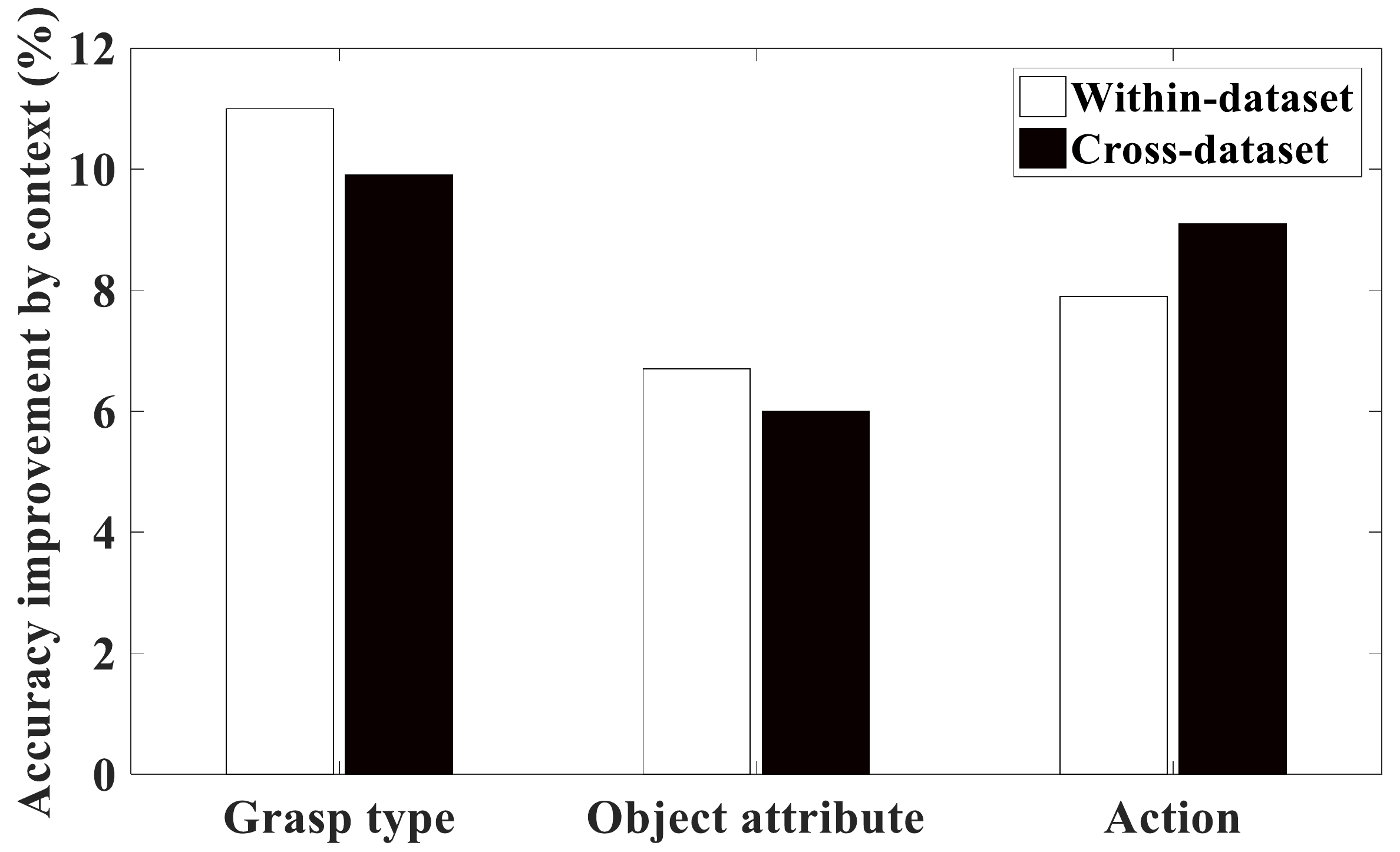}
    \caption{Generalizability evaluation of contextual relationship by classification accuracy improvement of grasp types, object attributes and actions. Experiments are conducted on GTEA Gaze Dataset with CNN classifiers trained on the same dataset. For ``within-dataset'', context learned on GTEA Gaze Dataset is used, while for ``cross-dataset'', context learned on GTEA Dataset is used.}
    \label{fig_cross_context}
\end{figure}

We then evaluate the cross-dataset performance of the contextual relationships studied in this work. Specifically, we are interested in how the context learned from one dataset (GTEA Dataset) could improve the recognition performance in another dataset (GTEA Gaze Dataset). Figure \ref{fig_cross_context} shows cross-dataset performance of context by the improvement of classification accuracy of grasp types, object attributes and actions respectively. Compared with the context learned on the current GTEA Gaze Dataset, the context learned on previous GTEA Dataset achieves comparative accuracy improvement (and even larger improvement for action classification), indicating good generalizability of contextual relationships to different datasets of HOM activities. More importantly, the generalizability study shows the importance of exploring common knowledge shared within the same problem domain in the realization of a more general vision recognition system.

\section{Conclusion and future work}
\label{s_conclusion}
In this work, we propose a novel method for understanding hand-object manipulation, in which grasp types, object attributes and actions are recognized in an unified framework. Considering various contextual relationships between actions, hands and objects, we construct a context model to optimize the three tasks together.

Experiments on public egocentric activity datasets illustrate that the proposed method achieves best recognition performance for all three tasks, verifying the effectiveness of jointly modeling grasp types, object attributes and action. The visualization and analysis of the learned context support our hypothesis that the contextual relationships between action, hand and object play an important role in understanding hand-object manipulation. 

In current work, the contextual relationships between actions, hands and objects are studied in a static model, in which the relationships remain unchanged during a manipulation action. However, in certain action such as opening a tight bottle cap, hand grasp type changes from the beginning (\textit{power sphere}) to the end (\textit{precision sphere}). One direction of our future work is to study the contextual relationships dynamically in order to capture the temporal evolution of hand-object interaction patterns in different manipulation actions. Besides, to train our model, we need to annotate lots of information in each training image, which limits the scale of data we can explore. Therefore, another direction of our future work is to develop an efficient approach (such as unsupervised learning) for understanding hand-object manipulation from a large amount of data.

\bibliographystyle{SageH}
\bibliography{reference}

\end{document}